\documentclass[a4paper, twocolumn, 8pt]{extarticle}

\pdfoutput=1

\usepackage[T1]{fontenc}
\usepackage[compact,small]{titlesec}
\usepackage[english]{babel}
\usepackage[margin=18mm, columnsep=7mm]{geometry}
\usepackage[pagebackref]{hyperref}
\usepackage[utf8]{inputenc}
\usepackage{abstract}
\usepackage{amsfonts}
\usepackage{amsmath}
\usepackage{amssymb}
\usepackage{amsthm}
\usepackage{authblk}
\usepackage{caption}
\usepackage{float}
\usepackage{graphicx}
\usepackage{lmodern}
\usepackage{pdfpages}
\usepackage{subfig}
\usepackage{subfiles}
\usepackage{upgreek}

\usepackage{cleveref}

\crefname{section}{Sec.}{Secs.}
\crefname{figure}{Fig.}{Figs.}
\crefname{equation}{Eq.}{Eqs.}

\hypersetup{
    bookmarksopenlevel=1,                                                                                                                                 
    pdftitle={Updating the generator in PPGN-h with gradients flowing through the encoder},                                                               
    pdfauthor={Hesam Pakdaman},                                                                                                                           
    pdfsubject={Deep Learning},                                                                                                                           
    pdfkeywords={Computer Science, Computer Vision, Deep Learning, Generative Adversarial Networks, Machine Learning, Neural Networks, Generative models},  
    colorlinks=true                                                                                                                                       
}

\theoremstyle{definition}

\theoremstyle{definition}

\theoremstyle{definition}

\providecommand{\pdata}{\ensuremath{{p_\text{data}}}}
\providecommand{\pgen}{\ensuremath{{p_g}}}
\providecommand{\fun}[3]{\ensuremath{#1\!: #2 \rightarrow #3}}

\providecommand{\lossgan}{\ensuremath{L_\text{gan}}}
\providecommand{\lossx}{\ensuremath{L_x}}

\providecommand{\lossh}{\ensuremath{L_{h}}}

\providecommand{\fc}[1]{\textsf{fc#1}}

\title{\Huge\textbf{Updating the generator in PPGN-$h$ with gradients flowing through the encoder}}

\author{Hesam Pakdaman}

\affil{30 credits second-cycle degree project \\ School of Electrical Engineering and Computer Science, \\
KTH Royal Institute of Technology \\ \texttt{hesamp@kth.se}}

\date{}

\begin{document}

\maketitle
\addcontentsline{toc}{section}{Contents}


\begin{abstract}  \noindent  The Generative  Adversarial  Network  framework  has shown  success  in
implicitly modeling data  distributions and is able to generate  realistic samples. Its architecture
is comprised of a generator, which produces fake  data that superficially seem to belong to the real
data  distribution, and  a discriminator  which is  to distinguish  fake from  genuine samples.  The
Noiseless Joint Plug \&  Play model offers an extension to the  framework by simultaneously training
autoencoders. This model  uses a pre-trained encoder  as a feature extractor,  feeding the generator
with global information. Using the Plug \& Play network as baseline, we design a new model by adding
discriminators to  the Plug  \& Play  architecture. These additional  discriminators are  trained to
discern real and fake latent codes, which are  the output of the encoder using genuine and generated
inputs,  respectively. We  proceed  to  investigate whether  this  approach  is viable.  Experiments
conducted for the MNIST manifold show that this indeed is the case. \end{abstract}

\section{Introduction}\label{sec:intro}

Generative models  can learn data  distributions, explicitly or  implicitly depending on  the model.
When machines are to better understand data, the  use of such models become relevant. From a socital
point of view the  advancement of generative models are deserving for they  will help with instances
where we are interested in the manifold itself, rather than predicting some quantity or class.

Generative Adversarial Network (GAN) \cite{goodfellow2014generative}  is a framework that implicitly
estimates a given  data distribution and has  sampling capabilities \cite{goodfellow2014generative}.
Its   application  include   image-to-image   translation   \cite{isola2016image},  generating   art
\cite{nguyen2016plug},  text-to-image  synthesis   \cite{reed2016generative}  and  visualization  of
learned  representations  \cite{dosovitskiy2016generating}.  Two  entities  are  integral  to  GANs.
The  \emph{generator}  that  tries  to  produce  samples  indistinguishable  from  those  pertaining
to  the   data  distribution   and  the   \emph{discriminator}  that  tries   to  tell   them  apart
\cite{goodfellow2014generative}.  These  two  entities  have conflicting  goals.  The  generator  is
to  trick  the   discriminator  by  presenting  samples   that  seem  to  come   from  the  dataset,
while  the   discriminator  improves  on  its   task  of  separating  generated   and  real  samples
\cite{goodfellow2014generative}.  This  creates an  adversarial  situation  in which  the  generator
wants  to  maximize  the  chances  that   the  discriminator  errs,  as  the  discriminator  combats
this  \cite{goodfellow2014generative}.  Ideally,  the  generator  better  estimates  the  true  data
distribution  as it  progresses in  its task  of tricking  the discriminator.  In the  original work
\cite{goodfellow2014generative}  the authors  present  an intuitive  explanation:  the generator  is
imagined  as a  forger trying  to slip  by an  inspector, the  discriminator, with  fake goods.  The
success of  the forger  stems from  how realistic the  products feel,  while the  inspector measures
accomplishment  with how  capable  it is  of  stopping  the lawbreaker.  This  framework has  gained
popularity in the  deep learning community due to the  fact that it can be used  in conjunction with
the backpropagation algorithm and for its efficient sampling capability.

Recent  works  have   introduced  the  idea  to   feed  the  generator  with  data   coming  from  a
higher-level layer of some pre-trained encoder network \cite{nguyen2016synthesizing, nguyen2016plug,
dosovitskiy2016generating}.  Since  these  features  reside   in  some  intermediate  layer  of  the
encoder  and   are  latent,  we   refer  to  them  as   hidden  representations  or   latent  codes.
Given  a  hidden  representation  produced  by  the   encoder,  the  generator  can  be  trained  to
reconstruct the  input to the  encoder that caused the  representation \cite{nguyen2016synthesizing,
dosovitskiy2016generating}. We  say that the  generator inverts a target  network, in this  case the
encoder.  High-level  codes  contain abstract  features  that  can  be  used by  the  generator  for
producing high-quality samples  \cite{nguyen2016synthesizing}. This can be better  understood in the
context  of  encoding  and  generating  images.  As  we  move  from  a  lower  to  higher  layer  in
the  encoder, the  features  go  from containing  local  information, as  in  edges  or corners,  to
more  abstract  representations  that  hold  global information,  e.g.  volume  or  object  category
\cite{nguyen2016synthesizing}. Images created from high-level codes, in contrast to lower-level, are
of greater  quality and  it is  hypothesized that  the generator fares  better when  it is  fed with
global  information \cite{nguyen2016synthesizing}.  This idea  is strengthened  by results  shown in
\cite{nguyen2016synthesizing}.

Stacked  Generative  Adversarial  Networks  (SGANs)  \cite{huang2016stacked}  use  several  encoders
and   generators   to   create   a   generative    model   based   on   the   GAN   framework,   see
\cref{fig:method:stackedgan}. The  authors bundle  generators together  to form  a stack,  with each
output  being the  input for  the subsequent  generator respectively  \cite{huang2016stacked}. Every
generator in  the stack  matches, dimensionwise, input  and output to  two successive  encoders. The
lower-placed encoder's output  is matched to the generator's output  and the higher-placed encoder's
output  with the  generator's input  \cite{huang2016stacked}. Similar  to the  generator stack,  all
encoders are  placed in line  to create an  encoder stack \cite{huang2016stacked}.  Furthermore, the
authors in  \cite{huang2016stacked} include additional  discriminators in the model.  These separate
the true hidden representations that an encoder outputs from those produced by a generator one level
higher.  This way  the  generator  is forced  to  match statistics  with  the  true hidden  manifold
\cite{huang2016stacked}.

In this master's thesis  we look at a particular generative model, the  Noiseless Joint Plug \& Play
Generative Network (PPGN) \cite{nguyen2016plug}, that combines the training of autoencoders with the
GAN framework. We  then investigate whether it  is viable to include  additional discriminators that
tell latent  codes in the  encoder space  apart, with the  potential benefit of  improving generated
samples and reducing model  complexity. The main difference with SGAN and  the method proposed here,
beside the architecture  and training algorithm used, is  that the output from the  generator is not
directly pitted against  the true latent code as in  \cite{huang2016stacked}. Instead, the generated
sample is pushed once  more through the encoder before handing it over  to a relevant discriminator,
see \cref{sec:method}  for a  more detailed description.  Nevertheless, we do  not claim  any method
better than the other. A thorough comparison between SGAN and Noiseless Joint PPGN, using the method
proposed here, is beyond the scope of this  master's thesis.

The main contribution of this thesis is  an investigation into the viability of attaching additional
discriminators to the architecture of Noiseless Joint  PPGN and an exposition of relevant generative
models. We provide a detailed background of  the generative framework and its relevant extensions in
\cref{sec:relwork}. Our  proposed network  design is  presented in  \cref{sec:method}, for  which we
conduct  and  showcase several  experiments  in  \cref{sec:exp}. The  results  are  compared to  the
Noiseless  Joint PPGN  in \cref{sec:discussion}  and a  discussion about  feasibility of  the method
ensues. Finally, in \cref{sec:con} we summarize our work.

\section{Background}\label{sec:relwork} First we  introduce the GAN framework for  which every model
described in  this section  incorporate. Thereafter,  we present two  extensions that  stabilize the
training procedure of GANs  and are used in our implementation. Next, the  PPGN and its predecessors
are  detailed in  chronological  order  and finally  we  present SGANs,  which  inspired  us to  use
additional discriminators.

The  GAN framework  introduced in  \cite{goodfellow2014generative} provides  a schematic  to develop
generative  models.  GANs consist  of  two  functions  $D, G$,  a  random  variable $z$  with  known
distribution and a value function $V(D,G)$  \cite{goodfellow2014generative}. The goal is to estimate
the  probability density  $\pdata$ of  a  given data  manifold $X$  \cite{goodfellow2014generative}.
Pushing   the  random   variable  through   the  generative   function  $\fun{G}{Z}{X}$   implicitly
specifies  a   distribution  $\pgen$,   which  with  parameter   tuning  should   approach  $\pdata$
\cite{goodfellow2014generative}. Convergence occurs when the  generator $G$ fully captures $\pdata$.
$D(x)$ is a  discriminative function that classifies its  input as real or fake,  where real samples
come from  the data  manifold $X$  and fakes are  generated by  $G$ \cite{goodfellow2014generative}.
The  generator has  to  trick the  discriminator  by  presenting samples  that  resemble real  ones,
in  effect  moving $\pgen$  towards  $\pdata$  \cite{goodfellow2014generative}. Simultaneously,  the
discriminator  will  try  to  tell  fakes  apart  from real  ones  to  negate  the  efforts  of  $G$
\cite{goodfellow2014generative}. This  creates an  adversarial situation that  can be  understood as
a  two-player  game,  wherein the  presence  of  a  discriminator  forces the  generator  to  better
estimate $\pdata$  if it wants  to succeed \cite{goodfellow2014generative}.  To this end,  the value
function  is  employed  for  which  it  is  necessary  that  maximization  of  it  yields  a  better
discriminator and minimization to a better generator \cite{goodfellow2014generative}. Therefore, the
adversarial  process  can  be summarized  as  a  two-player  minimax  game, $\min_G  \max_D  V(D,G)$
\cite{goodfellow2014generative}. Whenever the generator and the discriminator are unable to improve,
equilibrium is reached and the game  ends \cite{goodfellow2014generative}. We wish for the estimated
data distribution to tend towards the true distribution $\pgen \rightarrow \pdata$, meaning that the
discriminator will be unable to classify its input better than random. That is, as $\pgen$ converges
to $\pdata$ we have that $D(x) \rightarrow \frac{1}{2}$ \cite{goodfellow2014generative}.

In deep  learning scenarios, the  adversarial functions are set  as feedforward neural  networks and
optimization  of  the value  function  is  achieved with  gradient  descent,  respectively for  each
network  \cite{goodfellow2014generative}.  $V(D,G)$  in  \cite{goodfellow2014generative}  is  chosen
such  that  $G$  minimizes  the  Jensen-Shannon   divergence  (JSD)  between  $\pdata$  and  $\pgen$
\cite{goodfellow2014generative,  arjovsky2017towards},   while  $D$  maximizes   the  log-likelihood
$p(y|x)$ \cite{goodfellow2014generative}. Here $y$ is  a binary variable representing whether sample
point $x$  is fake or  real. In  the case of  \cite{goodfellow2014generative} the value  function is
given as
\begin{equation}
    \begin{aligned}
        \label{eq:jsd}
        \textstyle\min_G\max_DV(D,G)  &=  \mathbb{E}_{x  \sim  \pdata(x)}[\log  D(x)]\\
        &+\mathbb{E}_{z\sim p_z(z)}   [\log{(1-D(G(z)))}]
    \end{aligned}
\end{equation}
Note that when  using gradient descent the  discriminator influences the generator. This  due to the
fact that the gradient  $\nabla_{\theta}\mathbb{E}_{z\sim p_z(z)}[\log{(1-D(G(z)))}]$, which is used
for updating  $G$ with parameters  $\theta$, includes $D$. \cite{goodfellow2014generative}  To quote
the authors in \cite{goodfellow2014generative}, $G$ is  updated with ``gradients flowing through the
discriminator''.

Under the conditions in \cite{goodfellow2014generative},  the authors report problems with stability
in the training procedure of  GANs; 1) with the gradient for updating $G$  vanishing or 2) with mode
collapse of $\pgen$  when estimating multimodal $\pdata$. To address  the vanishing gradient problem
an  alternative value  function is  presented, $G$  is trained  to maximize  $\log{D(G(z))}$ instead
to  overcome  saturation  \cite{goodfellow2014generative}.  To  avoid  mode  collapse,  the  authors
\cite{goodfellow2014generative}  suggest training  $D$ and  $G$ asymmetrically,  specifically it  is
recommended  not  to  update  $G$  more  often than  $D$.  The  aforementioned  problems  have  been
theoretically investigated in \cite{arjovsky2017towards} and  practical remedies have been developed
\cite{arjovsky2017towards, salimans2016improved}. The  issues seem to be ameliorated  by choosing an
estimate of the Wasserstein metric instead of the JSD \cite{arjovsky2017wasserstein}
\begin{equation}
    \begin{aligned} 
        \label{eq:wassersteinestimate} 
        \textstyle  \max_{w\in \mathcal{W}} \mathbb{E}_{x\sim \pdata(x)}[f_w(x)]
                                            - \mathbb{E}_{z\sim   p(z)}[f_w(G(z))]  
    \end{aligned}   
\end{equation}   
where  $\{f_w\}_{w\in  \mathcal{W}}$ is  a  family  of  real-valued $K$-Lipschitz  scalar  functions
with  weight  space  $\mathcal{W}$.  The  quality  of estimating  the  true  Wasserstein  metric  is
dependent  of   the  choice   of  $\mathcal{W}$   \cite{arjovsky2017wasserstein}.\footnote{The  true
Wasserstein   metric  can   be  retrieved   from  the   Kantorovich-Rubinstein  duality   for  which
supremum  is  taken  over  all  real-valued,   scalar  $1$-Lipschitz  functions.  If  this  supremum
can  be found  for  some  $w\in\mathcal{W}$, then  \cref{eq:wassersteinestimate}  will  be within  a
constant  factor  from  the true  Wasserstein  value.  For  more  details  we refer  the  reader  to
\cite{arjovsky2017wasserstein}.} Practically  however, we  use a  discriminator\footnote{The authors
use the  term \emph{critic}  instead. This is  reminiscent of the  actor-critic terminology  used in
reinforcement  learning.  In  this  master's  thesis  we  adhere  to  the  original  formulation  in
\cite{goodfellow2014generative}.}  $D$  with weights  $\varphi$  in  lieu  of the  functions  $f_w$,
use  gradient ascent  to  find  $\varphi$ that  better  satisfies \cref{eq:wassersteinestimate}  and
try  to ensure  the  Lipschitz  constraint by  clamping  $\varphi$ to  be  within  some compact  box
after every  discriminator update  \cite{arjovsky2017wasserstein}. The generator  $G$ is  trained to
minimize \cref{eq:wassersteinestimate} with gradient descent $-\nabla_\theta\mathbb{E}_{z\sim p(z)}[
f_w(G(z))$], which  is similar to original  work in \cite{goodfellow2014generative}. The  authors in
\cite{arjovsky2017wasserstein}  encourages  training $D$  and  $G$  asymmetrically, ideally  $D$  is
trained more  for a better approximation  of \cref{eq:wassersteinestimate} given a  fixed $G$. Since
the discriminator is to estimate  the Wasserstein metric \cref{eq:wassersteinestimate} that requires
real-valued scalar functions, contrary to  \cite{goodfellow2014generative} $D$ no longer outputs the
probability of  real or fake. Thus,  the last softplus  activation of the original  discriminator in
\cite{goodfellow2014generative} is omitted \cite{arjovsky2017wasserstein}. Empirical results suggest
that  the  Wasserstein  metric  correlates  better  with  generated  sample  quality  than  the  JSD
\cite{arjovsky2017wasserstein}. In literature,  WGAN denotes that the Wasserstein metric  is used in
conjunction with the GAN framework.

In certain situations, the practice of  satisfying the $K$-Lipschitz constraint with weight clipping
has  adverse effects,  e.g. $\pgen$  not  converging or  $G$ displaying  poor sampling  capabilities
\cite{gulrajani2017improved}.  Instead, gradient  penalization  of $D$  helps  with convergence  and
allows for $G$  to produce higher-quality samples \cite{gulrajani2017improved}.  However, the method
is more  computationally demanding. The authors  base their idea  on the fact that  a differentiable
function is  $1$-Lipschitz if  it has gradient  norm of  at most  $1$ on the  whole domain  and vice
versa  \cite{gulrajani2017improved}.\footnote{If  $K\geq  1$,  then  an  $1$-Lipschitz  function  is
$K$-Lipschitz. This follows from the definition of Lipschitz continuity.} A lax approach would be to
penalize the  norm of  $D$, over  a smaller  set of  points, with  the distance  from $1$  using the
squared  Euclidean metric  \cite{gulrajani2017improved}. Therefore  in \cite{gulrajani2017improved},
the objective  function for  $D$ \cref{eq:wassersteinestimate} is  augmented by  $\textstyle \lambda
\mathbb{E}_{u\sim  p(u)}[(\|\nabla_u D(u)\|_2-1)^2]$,  where $\lambda$  is a  scalar to  control the
magnitude of the penalization and $u$ is a point  on the straight line between samples from true and
fake data  distributions \cite{gulrajani2017improved}.  Experiments, performed while  penalizing the
gradient of $D$  over a subset of points  $u$, is a good trade-off  between computational efficiency
and imposing  the constraint \cite{gulrajani2017improved}. The  authors \cite{gulrajani2017improved}
used  $u=\epsilon x  +  (1-\epsilon)\hat{x}$,  where $\epsilon$  followed  the uniform  distribution
$U[0,1]$. Results  in \cite{gulrajani2017improved} show an  improvement over WGAN. The  lax gradient
penalty approach applied to the WGAN framework is referred here to as WGAN-GP.

There are  works which experiment  with changing  the generator objective  in the GAN  framework. In
\cite{dosovitskiy2016generating}  it is  augmented by  a cost  that emanates  from some  layer in  a
network referred to as \emph{comparator}  $K$ \cite{dosovitskiy2016generating}. The intention is for
the generator  $G$ to  minimize perceptual  similarity between  the sample  it produces  and genuine
data  \cite{dosovitskiy2016generating}. Dissimilarity  is measured  with Euclidean  distance in  the
space  of  some  intermediate layer  of  $K$  and  this  provides  a  metric for  $G$  to  minimize.
Furthermore,  the  generator  is also  to  minimize  the  squared  distance between  fake  and  true
samples  to  match statistics  in  this  domain. Thus,  the  generator  objective is  expanded  with
perceptual  similarity loss  $\|K(G(z))-K(x)\|_2^2$  and a  loss  in $X$  space  $\|G(z) -  x\|_2^2$
\cite{dosovitskiy2016generating}.  Ablation studies  in  \cite{dosovitskiy2016generating} show  that
including  these two  $L_2$ losses  improve the  sampling  quality of  $G$. With  the new  generator
objective, the authors in \cite{dosovitskiy2016generating} visualize an encoder network, pre-trained
for image classification, using the GAN framework.  The hidden representation $h$ of an image, taken
from some layer of the encoder, is fed to  the generator which learns to reconstruct the image. This
can be seen as an  autoencoder-esque take on the GAN framework wherein $G$ takes  on the role of the
decoder that  is adversarially  trained \cite{nguyen2016synthesizing}. The  approach is  a departure
from the  original work  by \cite{goodfellow2014generative} where  the input to  the generator  is a
random variable $z$. The change means that $G$  loses its sampling capability and does not define an
implicit distribution  $\pgen$ \cite{nguyen2016synthesizing}. The  comparator $K$ can be  trained in
advance or  concurrent with the adversarial  functions $D, G$.  There are no restrictions  for which
task  $K$ is  trained  for, e.g.  the authors  \cite{dosovitskiy2016generating}  used a  pre-trained
classifier.

In \cite{nguyen2016synthesizing} the generator is trained to invert an encoder $E$ pre-trained on an
image manifold $X$,  following the methodology in \cite{dosovitskiy2016generating}  explained in the
previous paragraph.  The main contribution  of \cite{nguyen2016synthesizing}  is to use  the trained
generator  for visualizing  a  target network  $T$, utilizing  a  technique called  \emph{activation
maximization} \cite{erhan2009visualizing}. The  visualizing process consist of finding  an input $h$
that maximize a chosen output $c$ from some layer  in the target network $T_c$, such as a class unit
belonging to  a classifier. That  is, the search  is in  the domain space  of the generator  but the
objective is provided by the target network $T_c(G(h))$. Beginning with random input $h$, the latent
code is then iteratively optimized. The visualizations this technique yield are realistic, since $G$
acts as  a learned prior over  the data manifold that  the optimization process must  search through
\cite{nguyen2016synthesizing}. When  the target  is a  class output from  a classifier  network, $G$
is  able  to  produce high-quality  images.  However,  these  samples  lack diversity  as  shown  in
\cite{nguyen2016plug}  and the  reason is  that optimization  often leads  to the  same mode  of $h$
given  a fixed  target  unit \cite{nguyen2016plug}.  The  trained  generator $G$  can  also be  used
for  visualizing  other  networks  pre-trained  on  different  image  manifolds  with  good  results
\cite{nguyen2016synthesizing}.  The  insight  here  is  that  the  target  network  is  exchangeable
\cite{nguyen2016synthesizing, nguyen2016plug}. Since  the learned prior is a  deep generator network
that uses activation maximization, it is colloquially known as DGN-AM.

\begin{figure*}[t]
    \centering
    \subfloat[Noiseless Joint PPGN-$h$. Here  the encoder $E$ outputs a code $h$  which is the input
    to the generator $G$. We also see a  classifier $C$, used in the MALA-approx sampler, that takes
    generated images $\hat{x}$ and outputs labels. \label{fig:method:ppgn}]{
        \includegraphics[scale=0.5]{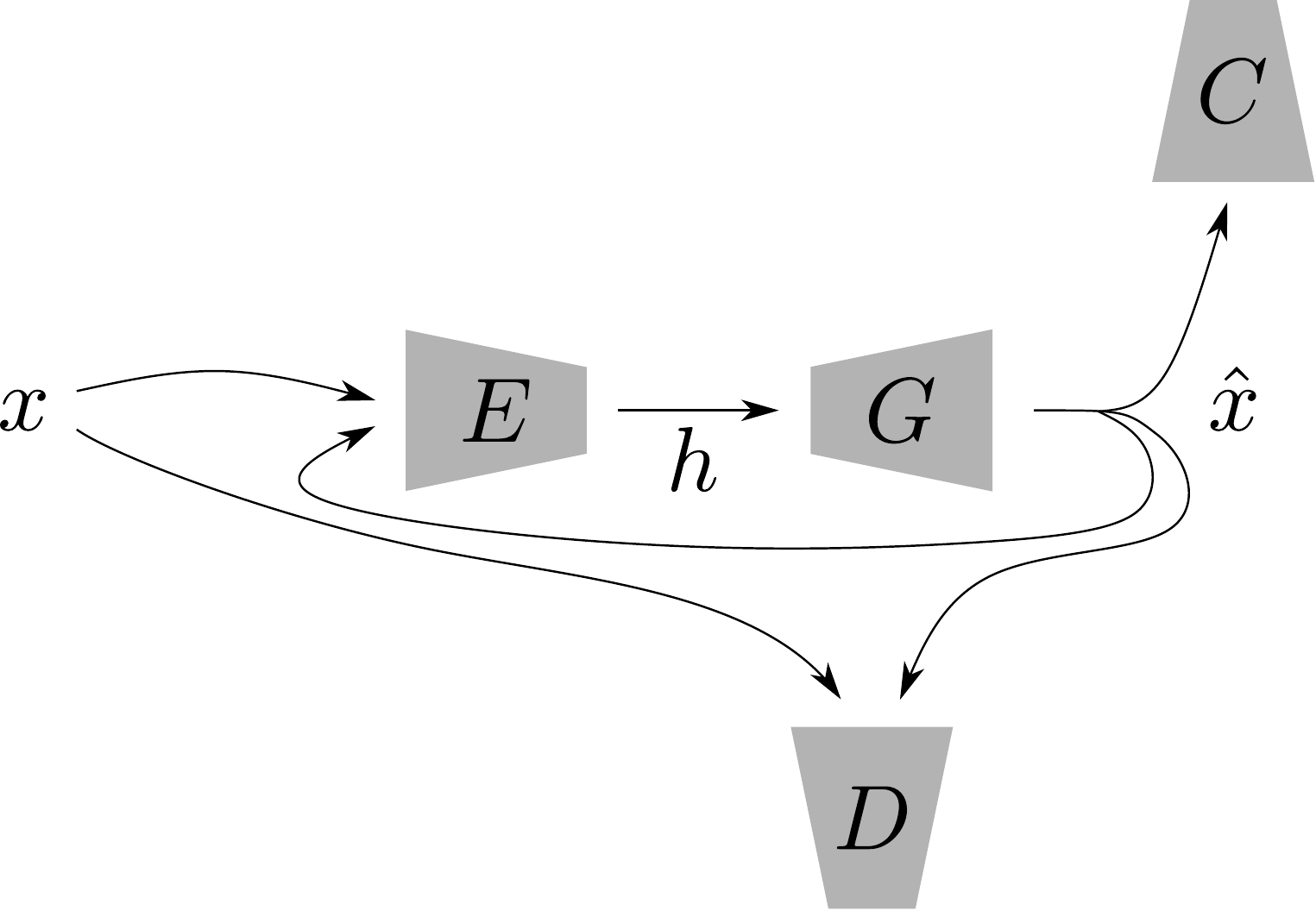}
    }\hspace{0.3cm}
    \subfloat[Proposed network  design. An additional  discriminator $D_i$  is attached to  the $i$th
    layer of encoder $E$. Dashed arrow indicate that both fake  and real samples propagate
    in the direction. \label{fig:method:design}]{
        \includegraphics[scale=0.5]{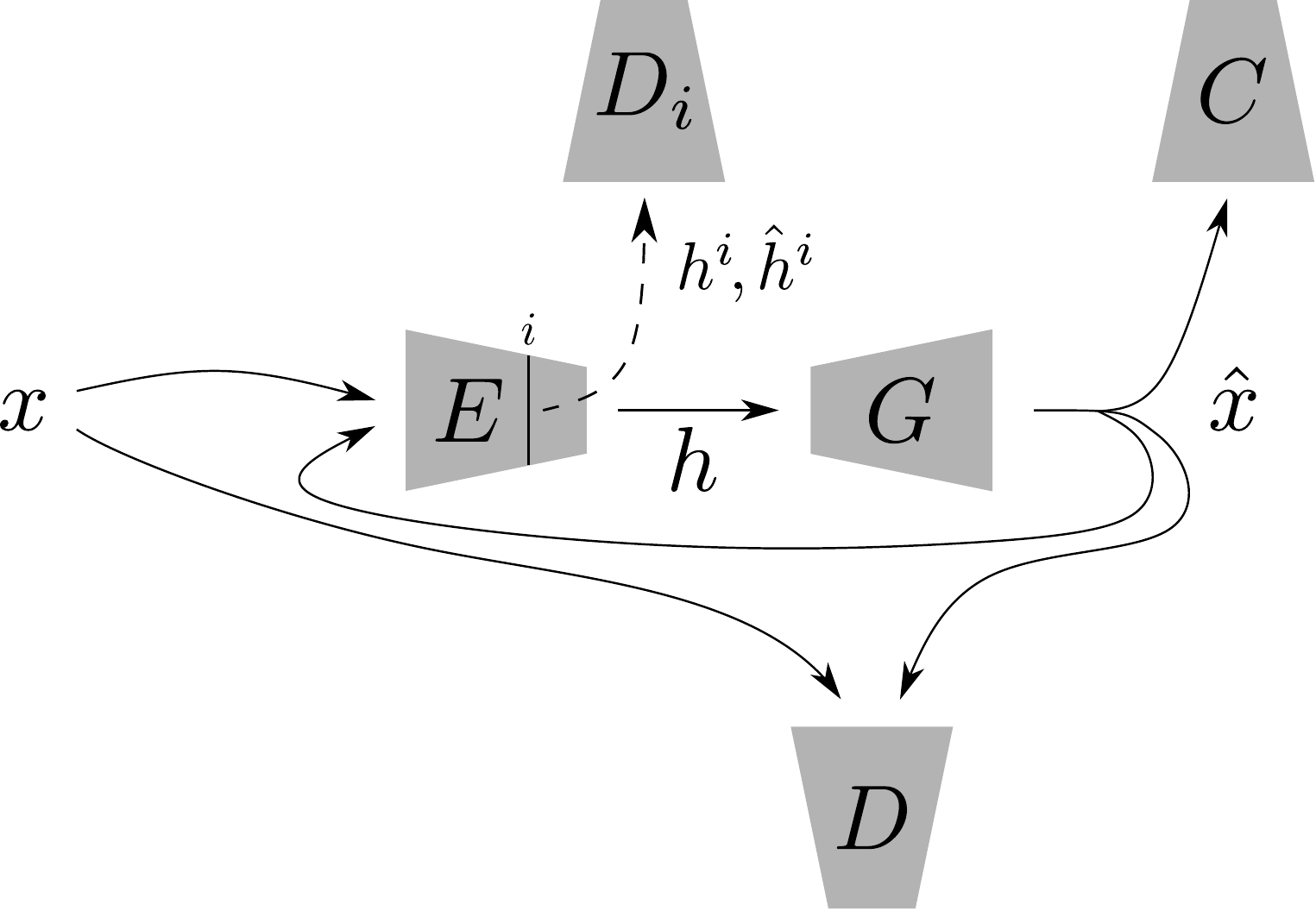}
    }

    \caption{Displaying   the  schematic   of   vanilla  PPGN   in   \ref{fig:method:ppgn}  and   in
    \ref{fig:method:design} our  design is shown.  We hypothesize that it  is possible to  train the
    generator $G$ while having attached discriminator $D_i$ to layer $i$ of encoder $E$.}

\end{figure*}

PPGN \cite{nguyen2016plug} is the successor to DGN-AM.  The authors characterize the DGN-AM model as
a joint  distribution $p(h,x,y)=p(h)p(x|h)p(y|x)$,  where $p(h)$  is a prior  over latent  codes $h$
produced by the encoder $E$, $G$ models  $p(x|h)$ and $p(y|x)$ is an exchangeable\footnote{Hence the
name Plug \& Play.} pre-trained network that classifies  targets $y$. Because $G$ does not define an
implicit  data  distribution  \cite{nguyen2016synthesizing} as  in  \cite{goodfellow2014generative},
given a latent code $h$ the fake variable $\hat{x}$ produced by $G$ is deterministic. Therefore, the
joint model can be written as  $p(h,y) = p(h)p(y|h)$ \cite{nguyen2016plug}. In \cite{nguyen2016plug}
they experiment with different formulations for the prior $p(h)$ with the aim of addressing problems
of sample  diversity and image quality  displayed by DGN-AM \cite{nguyen2016plug}.  They attain best
results by modeling  $h$ going through the  image space, in effect creating  a denoising autoencoder
via  the generator  $G$, $h  \rightarrow  \hat{x} \rightarrow  \hat{h}$ \cite{nguyen2016plug}.  This
specific model is called Joint PPGN-$h$ and  consists of four networks: a pre-trained encoder $E(x)$
that takes images $x$  as input, a generator $G(h)$ which has the  latent $h$-space produced by some
layer of $E$ as its domain, a discriminator  $D(x)$ capable of discerning fakes from real samples in
image space  $X$ and a pre-trained  classifier $C(x)$, see \cref{fig:method:ppgn}.  Drawing a sample
$x$ from the image manifold $X$ and feeding it to the composition $G(E(\cdot))$ implicitly give rise
to a  data distribution. The  GAN framework is  used to match with  the true data  distribution. The
generator $G$  is trained  using the loss  \begin{equation} \label{eq:meth:lg}\begin{aligned}  L_G =
\beta_1\lossx +  \beta_2\lossh + \beta_3\lossgan  \end{aligned} \end{equation} with  scaling factors
$\beta_k$, where $\lossx~=~||G(h) -  x||_2^2$ is an image loss, $\lossh~=~||\hat{h}  - h||_2^2$ is a
perceptual similarity  loss \cite{dosovitskiy2016generating}  and $\lossgan~=~-\log(D(G(h)))$  is an
adversarial loss. Here we denote the fake latent  code as $\hat{h} = E(G(h))$. Discriminator loss is
unaltered
    \begin{equation}
        \label{eq:method:lossdisc}
        \begin{aligned}
    L_D =  -\log D(x)-\log{(1-D(G(h)))}
        \end{aligned}
    \end{equation}
Sampling   is  achieved   iteratively  in   the  latent   code  space   of  $h$   using  a   derived
approximation\footnote{The   authors  ignored   the  reject   step  in   the  original   sampler  as
well  as   decoupled  the   $\epsilon_{12}$  and   $\epsilon_3$  terms   \cite{nguyen2016plug},  see
\cref{eq:first:mala}.} \cite{nguyen2016plug}  of the  Metropolis-adjusted Langevin  algorithm (MALA)
\cite{roberts1996exponential}, which is a Monte Carlo Markov Chain sampler. For some random variable
$v$ with probability distribution $p(v)$, the MALA-approx is written as
    \begin{align}\label{eq:first:mala} 
        v_{t+1} =  v_t  + \epsilon_{12} \nabla\log{p(v_t)} + N(0, \epsilon_3^2)
    \end{align} 
where $N(0,  \epsilon_3^2)$ is a sample  from the normal distribution  with variance $\epsilon_3^2$.
Using the MALA-approx, a sampler for the prior over $h$ conditioned on some class output $y_c$, i.e.
$p(h|y=y_c)$, can  be created \cite{nguyen2016plug}. Bayes'  rule tells us that  $p(h|y=y_c) \propto
p(h)p(y~=~y_c|h)$, which  used when  decoupling $\epsilon_{12}$  into $\epsilon_1$  and $\epsilon_2$
allows us to write \cite{nguyen2016plug}
    \begin{align}\label{eq:rel:mala}
        h_{t+1} &=  h_t  + \epsilon_1  \frac{\partial \log{p(h_t)}}{\partial h_t}
        + \epsilon_2  \frac{\partial  \log{p(y=y_c|h_t)}}{\partial h_t}\nonumber\\ &+ N(0, \epsilon_3^2) 
    \end{align} 
If  the prior  $p(h)$ is  modeled as  a denoising  autoencoder injected  with Gaussian  noise during
training, then
    \begin{align} 
        \frac{\partial \log{p(h)}}{\partial h} \approx \frac{R_h(h)-h}{\sigma^2}
    \end{align} 
assuming the variance of the noise  $\sigma^2$ is small \cite{nguyen2016plug}. Here $R_h(h)=E(G(h))$
is the reconstruction function  for the latent code $h$. Finally,  since the conditional probability
$p(y=y_c|h)$ is given by the classifier $C$, we simply write
    \begin{align} h_{t+1}\label{eq:rel:mala}
        &=  h_t  + \epsilon_1  (R_h(h_t)  -  h_t)  
        + \epsilon_2  \frac{\partial  \log{C_c(G(h_t))}}{\partial
        G(h_t)}\frac{\partial G(h_t)}{\partial h_t}\nonumber\\ &+ N(0, \epsilon_3^2) 
    \end{align} 
having used the chain rule on the $\epsilon_2$ term  to show that the gradient is forced through the
prior  over natural  images  given by  $G$  and  in addition  baked  $1/\sigma^2$ into  $\epsilon_1$
\cite{nguyen2016plug}.  $C_c$  denotes  the  output  of  class  unit  $c$  of  classifier  $C$.  The
parameters of  the MALA-approx \cref{eq:rel:mala}  each control  different aspects of  the generated
code.  The $\epsilon_1$  term encourages  the subsequent  generated code  $h_{t+1}$ to  be close  to
the  hidden  manifold  of  $h$  disregarding  any  class, seeing  as  $h$  will  be  guided  by  the
derivative of the denoising autoencoder \cite{nguyen2016plug}. Increasing the $\epsilon_2$ term will
construct codes that  are more favorable to  the classifier $C$ and  $\epsilon_3$ enforces diversity
\cite{nguyen2016plug}. In an experiment  the authors train the prior $p(h)$  of Joint PPGN-$h$ using
no  noise at  all, in  effect creating  a noiseless  autoencoder\footnote{This model  uses the  same
sampler found  in \cref{eq:rel:mala},  even though  the prior  $p(h)$ is  not trained  with Gaussian
noise.}. This  new model is called  Noiseless Join PPGN-$h$ and  achieves best results in  the paper
compared  to all  other PPGN  variants.  Noiseless Joint  PPGN produces  high-quality image  samples
with  resolutions  of $227  \times  227$  for all  $1000$  classes  in the  \mbox{ImageNet}  dataset
\cite{russakovsky2014imagenet}.

\begin{figure}[t]
    \centering
    \includegraphics[scale=0.80]{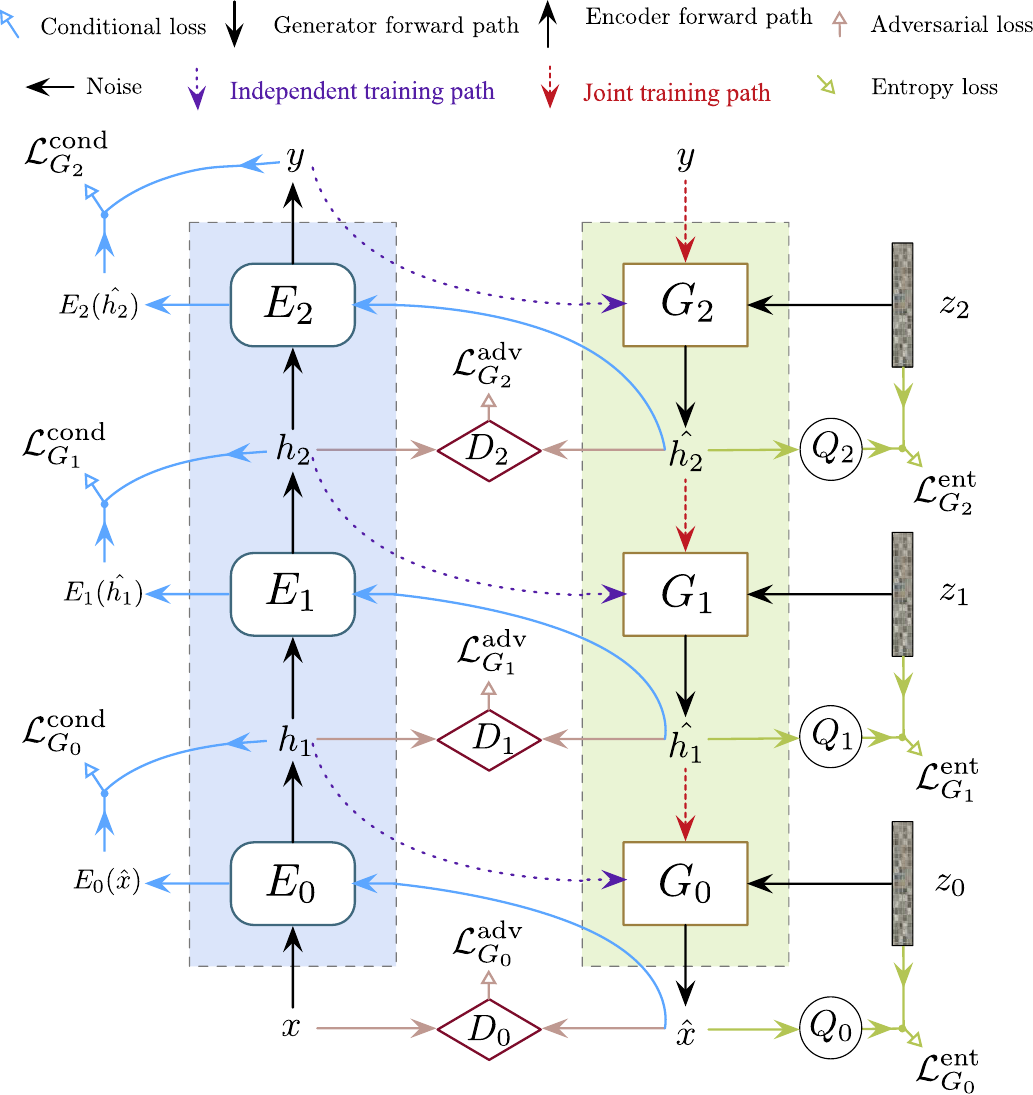}

    \caption{An overview of SGAN depicting a 3-stack  model, with the bottom-up encoder stack $E$ in
    the left box and the  top-to-bottom generator stack $G$ to the right. In  this case we see stack
    $G$  being trained  jointly.  The  ordering of  the  stacks refers  to  the  direction of  their
    respective input. Image taken from \cite{huang2016stacked}  and adjusted to fit a smaller frame.
    \label{fig:method:stackedgan}}

\end{figure}

Concurrent  to PPGN  \cite{nguyen2016plug} is  SGAN \cite{huang2016stacked},  which extends  the GAN
framework  \cite{goodfellow2014generative} by  creating  a  stack $G$  of  $N$  generators $G_i$  in
a  top-to-bottom  fashion.  The  idea  is  for  $G$  to  invert  a  pre-trained  stack  $E$  of  $N$
bottom-up  encoders  $E_i$.  The  ordering  of  the  stacks,  top-to-bottom  or  bottom-up,  reflect
the  direction  of  their  respective   input,  see  \cref{fig:method:stackedgan}.  Every  generator
$G_i$  is  fed  with  a  latent  code  $h_{i+1}$   coming  from  an  encoder  $E_i$  in  stack  $E$,
injected  with a  noise vector  $z_i$  and produces  feature  $\hat{h}_i =  G_i(h_{i+1}, z_i)$.  The
generated  $\hat{h}_i$ is  matched with  the input  to  encoder $E_{i}$.  Note that  this imposes  a
constriction  for  creating  the stacks,  the  input/output  pairs  of  the encoder  $E_i$  must  be
correctly  aligned  with  the generator  $G_i$.  Training  the  generative  networks is  first  done
independently  and  then  jointly  by  stacking them.  Therefore,  $\hat{h}_i  =  G_i(\hat{h}_{i+1},
z_i)$  when jointly  trained  and $\hat{h}_i  =  G_i(h_{i+1},  z_i)$ otherwise.  The  loss for  each
generator is  $\mathcal{L}_{G_i} = \mathcal{L}^{\text{adv}}_{G_i}  + \mathcal{L}^{\text{cond}}_{G_i}
+   \mathcal{L}^{\text{ent}}_{G_i}$.   The   adversarial  loss   $\mathcal{L}_{G_i}^{\text{adv}}   =
-\log{D_i(G_i(h_{i+1}, z_i))}$  is the same  as in \cite{goodfellow2014generative} but  modified for
hidden  representations.  $\mathcal{L}^{\text{cond}}_{G_i}  =  f[E_i(G_i(h_{i+1},  z_i)),  h_{i+1}]$
is  a  conditional loss  with  some  metric  $f$  (such as  the  $L_2$  norm  for latent  codes  and
cross  entropy  for  object categories)  introduced  to  assure  $G_i$  uses the  conditional  input
$h_{i+1}$. Lastly  $\mathcal{L}^{\text{ent}}_{G_i} =  -\log(Q_i(z_i|\hat{h}_i))$, where $Q_i$  is an
auxiliary  distribution for  the true  posterior $P_i(z_i,  \hat{h}_i)$ and  practically implemented
as  a  feedforward  network.  Since minimizing  $\mathcal{L}^{\text{ent}}_{G_i}$  is  tantamount  to
maximizing  a   variational  lower  bound   for  the  conditional   entropy  $H(\hat{h}_i|h_{i+1})$,
$\mathcal{L}^{\text{ent}}_{G_i}$ assures diversity of $\hat{h}_i$ by making the input $z_i$ relevant
for  $G_i$  when constructing  $\hat{h}_i$  \cite{huang2016stacked}.  In  addition to  the  ordinary
discriminator  inherent to  the GAN  framework,  with every  pair  $(E_{i}, G_i)$  of encoder  and
generator, a new discriminator $D_i$ is introduced  that is trained adversarially to tell the hidden
representations  $(h_i, \hat{h}_i)$  apart.  The discriminators  are trained  to  minimize the  loss
$\mathcal{L}_{D_i}  = -\log(D_i(h_i))  - \log(1-  D_i(G_i(h_{i+1}, z_i)))$  which is  similar as  in
\cite{goodfellow2014generative} but altered for taking  hidden representations as input. Samples are
produced by  conditioning the  top generator in  stack $G$  with label $y$  and injecting  its noise
vector.  The output  is then  fed to  the next  generator in  the stack  along with  the next  noise
vector.  This process  is  repeated  until $G_0$  is  reached, which  outputs  a  sample $\hat{x}  =
\hat{h}_0$. SGANs produce  high-quality samples for the MNIST  \cite{lecun1998gradient} and CIFAR-10
\cite{krizhevsky2009learning} datasets.

SGANs introduce the idea to discriminate  between hidden representations $(h_i, \hat{h}_i)$. In this
master's  thesis we  will use  this approach  for the  Noiseless Joint  PPGN-$h$ model,  but instead
discriminate  between pairs  of codes  produced  entirely by  the encoder.  As  we will  see in  the
following section, this  will not impose the  constraint abided by SGANs of  aligning each generator
layer with its corresponding encoder layer.

\section{Method}\label{sec:method}   We   start   with   the  noiseless   variant   of   the   Joint
PPGN-$h$\footnote{For brevity  we hereby  refer to  this model,  interchangeably, as  PPGN-$h$.} and
change its architecture such  that it includes new discriminators. Higher  layers in the pre-trained
encoder $E$  contain abstract features  with a space  that is smaller  in dimension compared  to the
lower layers.  \emph{We hypothesize that  it possible  to train the  generator $G$ in  PPGN-$h$ with
gradients flowing  through discriminators that are  attached in these compressed,  abstract spaces}.
For layer  $i$ in encoder  $E$ we attach  a discriminator $D_i$ to  discern fake codes  $\hat{h^i} =
E^i(G(h))\footnote{Here we denote the $j$th layer of  any network $y=A(x)$ with $A^j$ and its output
as $y^j$.}$ from real ones $h^i$ in the associated latent space. We augment the adversarial loss for
$G$\footnote{In this section we  present the loss functions without taking  any consideration to the
Wasserstein metric. Of course,  the necessary changes are simple to make, but  we chose the original
formulation as in PPGN-$h$ so that it is easier for the reader to compare with our extension.}
\begin{equation}
    \begin{aligned}
        \label{eq:method:lossgan}
        \textstyle \lossgan = -\lambda_0 \log{D(\hat{x})}  - \sum_j  \lambda_j \log{D_j(\hat{h}^j)}
    \end{aligned}
\end{equation}
where $\lambda_j$ are scaling factors and $D$ is the ordinary discriminator  in PPGN-$h$. If we set
$\lambda_0=1$ and  $\lambda_j = 0$ for  $j>0$, then we get  back the usual adversarial  loss for the
generator in  PPGN-$h$. Since there  is no restriction  in \cite{nguyen2016plug} for  choosing which
layers of $E$ to measure $L_2$ perceptual losses from, we can adopt the following simple policy. For
every discriminator  $D_j$ we attach,  we also include the  autoencoder reconstruction loss  of each
respective layer
\begin{equation}
    \begin{aligned}
        \label{eq:method:aeloss}
        \textstyle L_h =  \sum_j \alpha_j  ||\hat{h}^j - h^j||_2^2
    \end{aligned}
\end{equation}
where we have  scaling numbers $\alpha_j$. The  discriminator attached to the latent  space of $E^i$
use the loss
\begin{equation}
    \begin{aligned}
        L_{D_i} =  -\log D_i(h^i)-\log{(1-D_i(\hat{h}^i))}
    \end{aligned}
\end{equation}
which is similar to \cref{eq:method:lossdisc}.

Observe the difference  between the discriminators defined  in this master's thesis and  in the SGAN
paper. Here  we tell  autoencoder reconstructions $\hat{h}^i  = E^i(G(h))$ apart  from codes  $h^i =
E^i(x)$.  In SGAN  hidden fake  outputs from  $G_i$ is  directly compared  with real  inputs to  the
corresponding  encoder  network  $E_i$,  i.e.  $\hat{h}_i =  G_i(h_{i+1},  z_i)$  against  $h_{i}  =
E_{i-1}(x)$, see \cref{fig:method:stackedgan} and  \cref{fig:method:design}. Since fake hidden codes
$\hat{h}^i$ are produced  entirely by encoder $E^i$ using  fake inputs $\hat{x}$, we do  not need to
align input/output pairs of encoder and generator  layers. Thus, we skip the alignment constraint of
SGANs.

\begin{figure*}[t]
    \centering
    \subfloat[Vanilla PPGN-$h$. \label{fig:exp:vanilla}]{
        \includegraphics[scale=0.4]{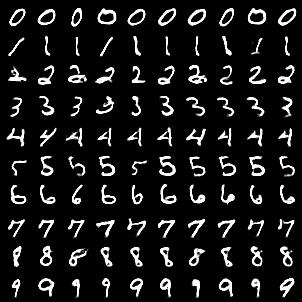}
    }\hspace{0.1cm}
    \subfloat[PPGN-$h$-$D_\fc1$ using \lossgan. \label{fig:exp:fc1gan}]{
        \includegraphics[scale=0.4]{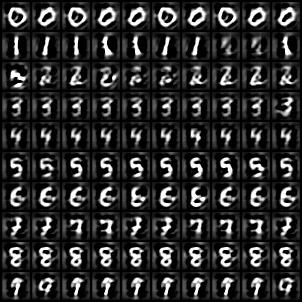}
    }\hspace{0.1cm}
    \subfloat[PPGN-$h$-$D_\fc1$ with full loss. \label{fig:exp:fc1full}]{
        \includegraphics[scale=0.4]{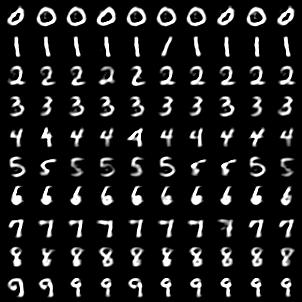}
    }

    \subfloat[PPGN-$h$-$D_\fc1$ using \lossh\ and \lossx. \label{fig:exp:fc1wo}]{
        \includegraphics[scale=0.4]{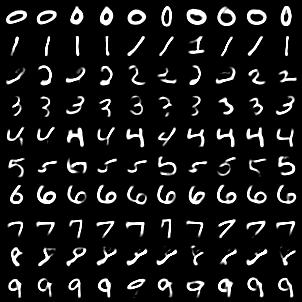}
    }
    \hspace{0.1cm}
    \subfloat[PPGN-$h$-combined. \label{fig:exp:combined}]{
        \includegraphics[scale=0.4]{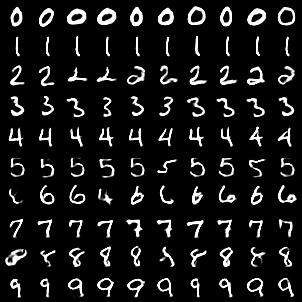}
    }
    \hspace{0.1cm}
    \subfloat[PPGN-$h$-random. \label{fig:exp:randomized}]{
        \includegraphics[scale=0.4]{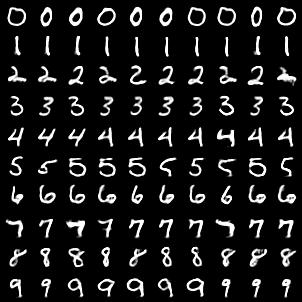}
    }

    \caption{In each figure the samples are produced by the same generator $G$ with fix architecture
    but trained differently. The parameters of MALA-approx, number of epochs and the scaling factors
    of  each partial  loss are  the  same across  every experiment.  \textbf{\ref{fig:exp:vanilla}}:
    We   begin  with   the   baseline  model   Vanilla  PPGN-$h$   following   the  methodology   in
    \cite{nguyen2016plug}. We can see some diversity within  each class and the generated digits are
    similar to  the MNIST  dataset that can  be seen in  \cref{fig:exp:compare:mnist}. We  note that
    class $7$ shows least diversity. \textbf{\ref{fig:exp:fc1gan}}: We replace the discriminator $D$
    in Vanilla PPGN-$h$ with $D_{\fc{1}}$ to create the PPGN-$h$-$D_\fc{1}$ model. For this specific
    experiment we  train $G$  using only the  adversarial loss \lossgan\  to see  if $G$ is  able to
    converge using  only gradients  that flow  through $D_\fc{1}$. The  results degrade  compared to
    baseline.  Nevertheless, $G$  has learned  shapes of  every  digit class  and we  can hint  some
    diversity for  numbers $0$  and $1$. Evidently,  this experiment shows  that only  including the
    adversarial  loss  \lossgan\ with  $D_\fc{1}$  is  not  sufficient for  generating  high-quality
    samples. \textbf{\ref{fig:exp:fc1full}}:  We train PPGN-$h$-$D_\fc{1}$  with full loss,  i.e. in
    addition \lossgan\  we also include  image loss \lossx\  and perceptual similarity  loss \lossh.
    Sampling  quality  improves  but is  not  quite  on  a  par  with baseline.  In  particular,  we
    notice  that  the  samples do  not  look  as  sharp  as  those generated  by  Vanilla  PPGN-$h$.
    \textbf{\ref{fig:exp:fc1wo}}:  Including  losses  \lossx  and  \lossh\  may  have  rendered  the
    adversarial  loss useless.  Given this  possibility,  we train  PPGN-$h$-$D_\fc{1}$ but  exclude
    \lossgan. The  samples are of  mixed quality, where  some digits, such as  $0, 1, 9$,  look good
    relative to baseline while for example $2, 4,  8$ seem worse. This suggests that the adversarial
    loss has had an impact on the  results in \ref{fig:exp:fc1full} since the samples look different
    from those produced in this experiment. However, the results for this model are sharper than the
    samples in \ref{fig:exp:fc1full}. \textbf{\ref{fig:exp:combined}}: A new model PPGN-$h$-combined
    is created by adding the discriminator $D_\fc{1}$ to Vanilla PPGN-$h$. We are interested to know
    whether  the model  is  able to  converge  when having  two  different discriminative  objective
    functions  for $G$  to minimize.  Generated samples  are of  good quality  and is  comparable to
    baseline  model.  Nonetheless,  the  training  time  of  this  model  was  considerably  longer.
    \textbf{\ref{fig:exp:randomized}}: The  PPGN-$h$-combined approach  showed good results  but had
    longer training time than Vanilla PPGN-$h$. To battle this, we randomized with equal probability
    which discriminator objective  and corresponding adversarial loss to optimize.  We refer to this
    model as PPGN-$h$-random.  The approach lead to  a reduction in training  time while maintaining
    the higher-quality sampling capability of PPGN-$h$-combined.}

\end{figure*}

\begin{figure*}[t]
    \centering
    \subfloat[Vanilla PPGN-$h$. \label{fig:exp:plot:vanilla}]{
        \includegraphics[scale=0.36]{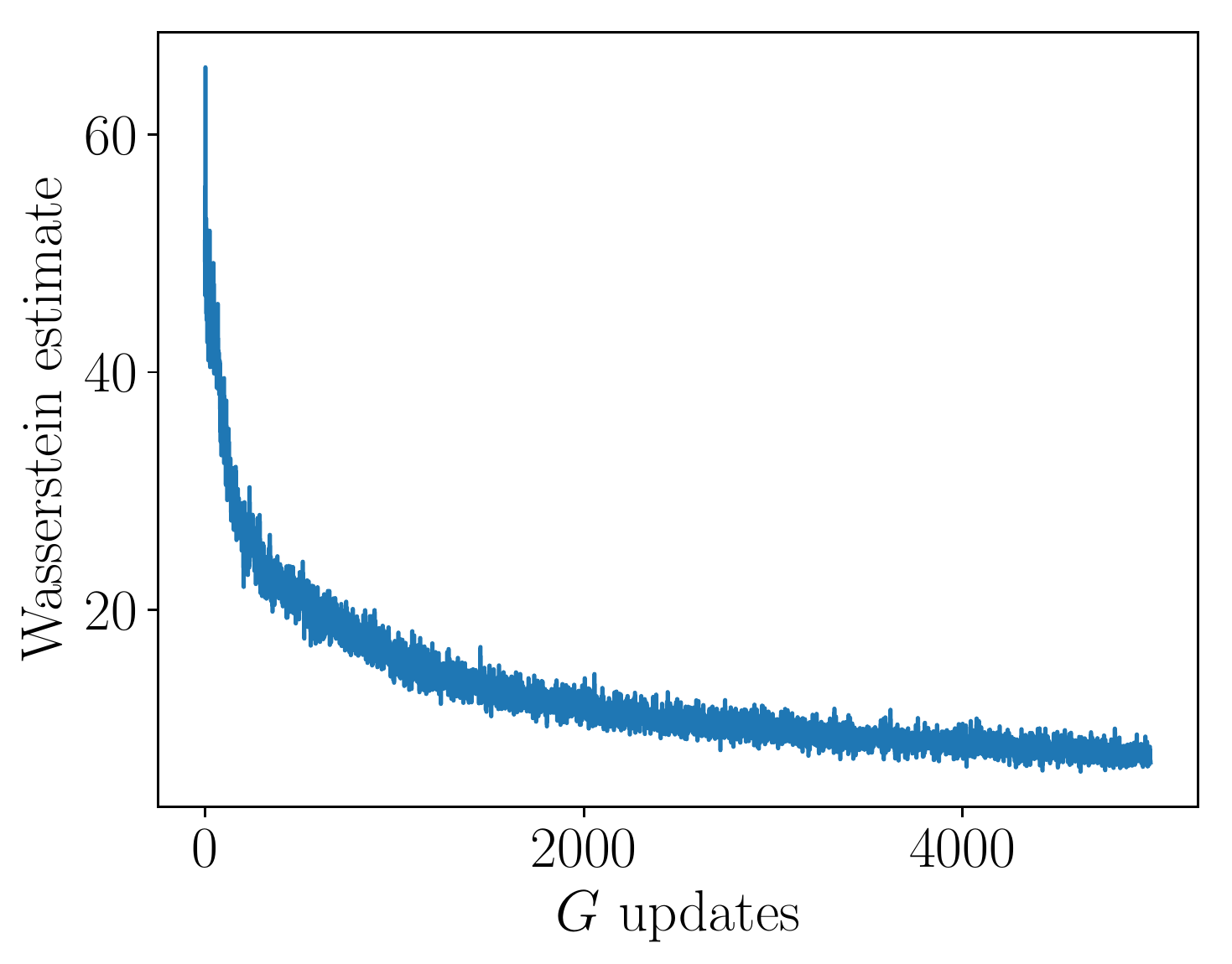}
    }\hspace{0.0cm}
    \subfloat[PPGN-$h$-$D_\fc1$ with full loss. \label{fig:exp:plot:fc1full}]{
        \includegraphics[scale=0.36]{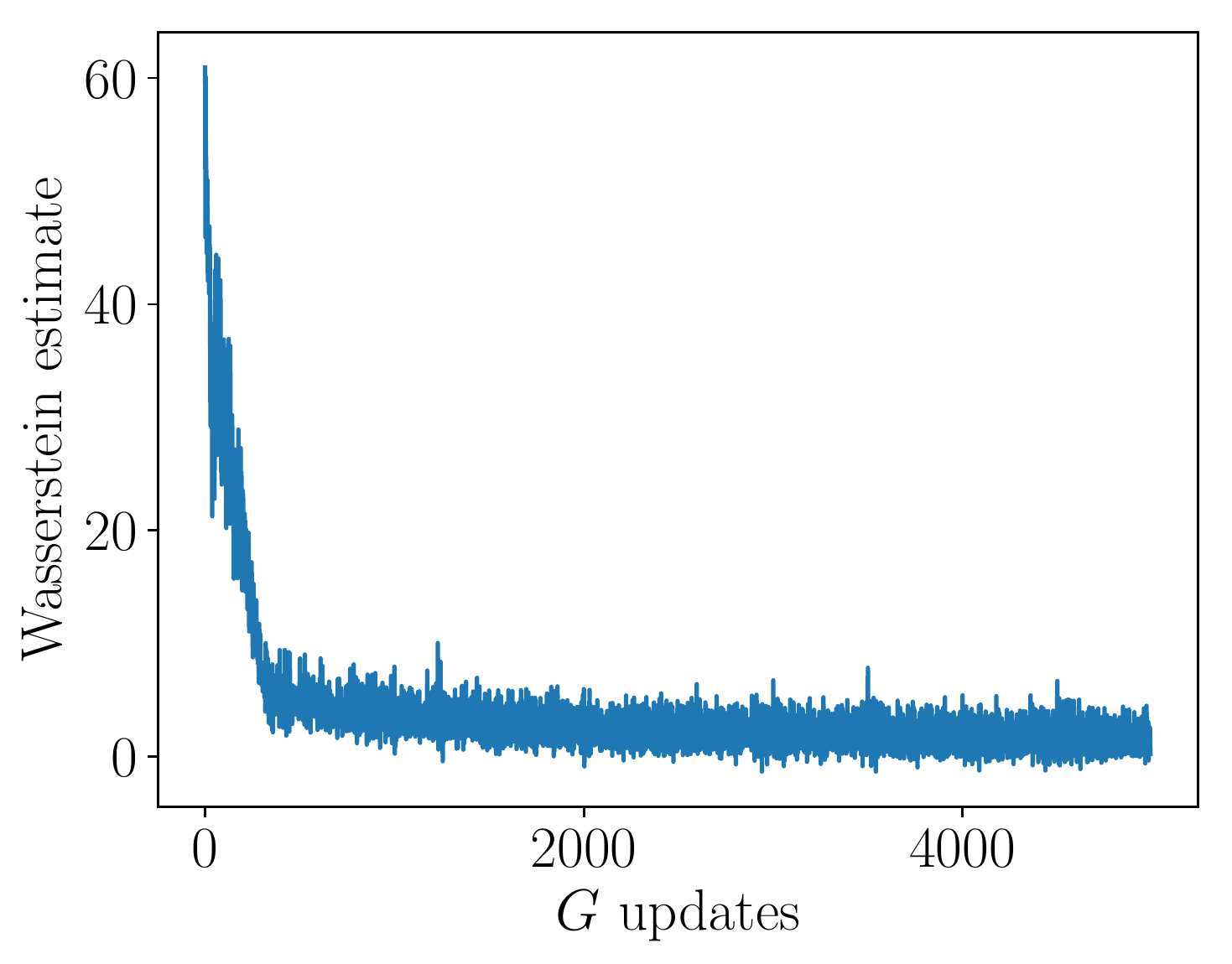}
    }
    \hspace{0.0cm}
    \subfloat[PPGN-$h$-combined. \label{fig:exp:plot:combined}]{
        \includegraphics[scale=0.36]{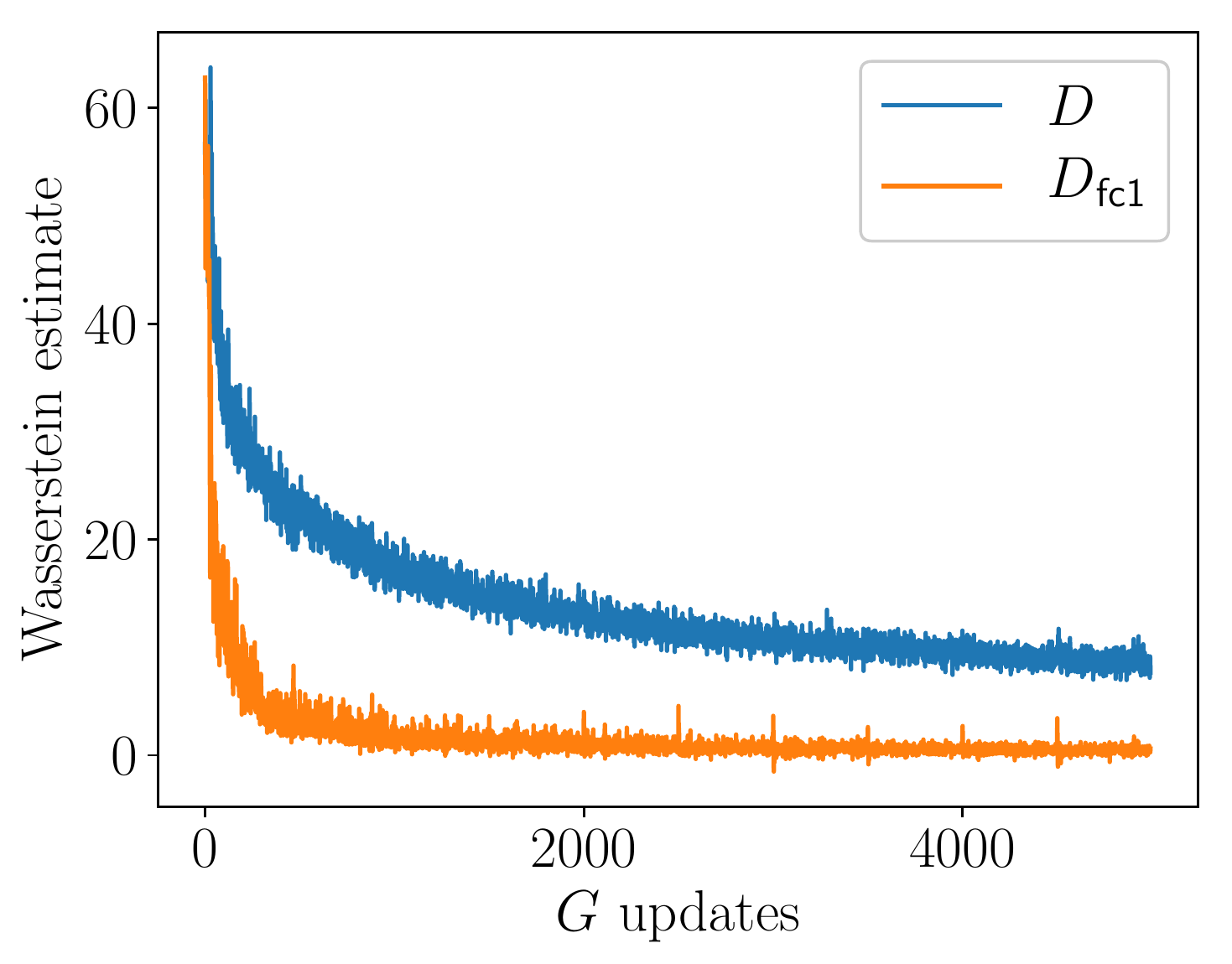}
    }

    \caption{We  chose  three models  and  plot  the  Wasserstein  estimate during  training.  Every
    model  was able  to minimize  the metric.  The Wasserstein  estimate provided  by $D_\fc{1}$  in
    \cref{fig:exp:plot:fc1full} seem easier for $G$ to  minimize since it flattens quickly near zero
    compared to  \cref{fig:exp:plot:vanilla}. A reason  might be  that $D_\fc{1}$ has  less capacity
    relative  to  $G$.  When including  both  discriminators  $D$  and  $D_\fc{1}$, we  can  see  in
    \cref{fig:exp:plot:combined} that both Wasserstein estimates are minimized by $G$.}

\end{figure*}

\begin{figure*}[t]
    \centering
    \subfloat[MNIST dataset. \label{fig:exp:compare:mnist}]{
        \includegraphics[scale=0.4]{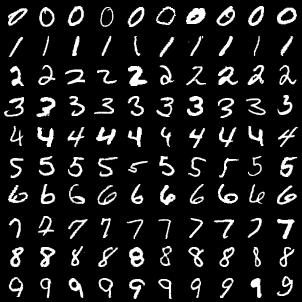}
    }\hspace{0.1cm}
    \subfloat[Baseline PPGN-$h$. \label{fig:exp:compare:vanilla}]{
        \includegraphics[scale=0.4]{./figures/experiment/vanilla}
    }\hspace{0.1cm}
    \subfloat[PPGN-$h$-random. \label{fig:exp:compare:randomized}]{
        \includegraphics[scale=0.4]{./figures/experiment/randomized}
    }

    \caption{To  facilitate  a  comparison  between  the  MNIST  dataset,  the  baseline  model  and
    PPGN-$h$-random  we showcase  them all  here.  Both of  these  two models  produce samples  that
    resemble MNIST digits, however they do not show as much diversity.}

\end{figure*}

\section{Experiments\label{sec:exp}}           We           use           freely           available
code\footnote{\url{github.com/caogang/wgan-gp}}        of        a       WGAN-GP        architecture
\cite{salimans2016improved}  as a  base for  implementing the  Noiseless Joint  PPGN-$h$ model.  The
WGAN-GP  algorithm trains  $G$ and  $D$ asymmetrically,  specifically the  discriminator is  trained
$100$  times for  the first  $25$ training  iterations  and the  same amount  every $500$th.  Unless
otherwise  stated,  we  use  the  same hyperparameters  as  in  \cite{salimans2016improved}.  For  a
full  overview  of  hyperparameters  and  for reproducibility  proposes,  we  publicly  release  the
code\footnote{\url{github.com/hesampakdaman/ppgn-disc}} used for this master's thesis.

In every experiment  we use the MNIST  dataset \cite{lecun1998gradient}, for which  the training set
contains  $60000$ handwritten  digits. Each  of these  images consist  of $28\times  28$ pixels  and
depicts a number between $0$ and $9$, see \cref{fig:exp:compare:mnist}. We normalize the data before
feeding it to the entire network.

The   encoder  $E$   is  a   convolutional  network\footnote{A   convolutional  or   deconvolutional
layer  associated  with  triplet  $(a,b,c)$  has  $a$  input  channels,  $b$  outputs  channels  and
a  kernel  size  of  $c\times  c$.  Both  of  these layer  types  use  a  stride  of  $1$.  A  fully
connected  layer  with  tuple  $(a,b)$  has  $a$   input  channels  and  $b$  output  channels.  All
max  pooling   layers  use  a   kernel  size  of  $2\times   2$.}  taken  from   freely  distributed
code\footnote{\url{github.com/pytorch/examples/tree/master/mnist}}.   However,   we   modified   the
architecture  and  ended  up  with  \textsf{conv1}$(1,  64,  7)$  $\rightarrow$  \textsf{conv2}$(64,
128,  7)$ $\rightarrow$  \textsf{pool2}  $\rightarrow$ \textsf{conv3}$(128,  256, 7)$  $\rightarrow$
\textsf{pool3}  $\rightarrow$ \textsf{fc1}$(256,  64)$  $\rightarrow$ \textsf{fc2}$(64,10)$.  Notice
that the last  two layers of $E$,  the fully connected layers \textsf{fc1}  and \textsf{fc2}, output
vectors of size $64$ and $10$ respectively. Every  convolutional and fully connected layer of $E$ is
followed by the ReLU  activation function, except for \fc{2}. We get the  classifier $C$ by applying
softmax to $\fc{2}$. $E$ is pre-trained for classification on MNIST using cross entropy loss and its
parameters  are held  fixed throughout  every  experiment. The  generator $G$  is a  deconvolutional
network  \cite{dosovitskiylearning}  \textsf{gen-fc1}\footnote{We  prefix any  newly  defined  fully
connected layer with an identifier since $\fc{1}$ and \fc{2} are reserved for the encoder $E$.}$(64,
1600)$  $\rightarrow$  \textsf{deconv2}$(64,  512, 5)$  $\rightarrow$  \textsf{deconv3}$(512,256,5)$
$\rightarrow$  \textsf{deconv4}$(256,256,7)$ $\rightarrow$  \textsf{deconv5}$(256, 1,  10)$ and  its
architecture is  held fix. The  deconvolutional and fully  connected layers of  $G$ make use  of the
ReLU  function, with  the  exception of  \textsf{deconv5}  which uses  no  activation function.  All
adversarial networks, including  discriminators to be defined, are trained  using the ADAM optimizer
\cite{kingma2014adam}, while the encoder $E$ uses the SGD optimizer.

\subsection{Vanilla PPGN-$h$\label{sec:exp:mnist:vanilla}}  We train PPGN-$h$ using  the same losses
in \cite{nguyen2016plug},  where $(\beta_1, \beta_2,  \beta_3)$ were set  to $(1, 10^{-1},  2)$ such
that every partial loss has the same order of magnitude when training commences. For this experiment
we stopped  after $15$ epochs  of training  (roughly $5000$ $G$  updates) using minibatches  of size
$32$.  For  the MALA-approx  sampler  we  use  parameters  $(\epsilon_1, \epsilon_2,  \epsilon_3)  =
(10^{-2},1,10^{-15})$ for  $200$ iterations. These  values were based on  \cite{nguyen2016plug}, but
we  have  increased  the $\epsilon_1$  factor  to  get  more  generic codes,  which  yielded  better
samples. In  addition, we  increased $\epsilon_3$  for more  diversity. The  number of  epochs, size
of  minibatch,  parameters  of MALA-approx  sampler  and  the  values  of $\beta_k$  are  fixed  for
every  subsequent experiment.  $D$  is a  CNN with  the  following architecture,  \textsf{conv1}$(1,
256,  3)$ $\rightarrow$  \textsf{conv2}$(256,  256, 3)$  $\rightarrow$ \textsf{pool2}  $\rightarrow$
\textsf{conv3}$(256, 256,  3)$ $\rightarrow$ \textsf{pool3} $\rightarrow$  \textsf{conv4}$(256, 512,
3)$ $\rightarrow$  \textsf{pool4} $\rightarrow$ \textsf{disc-fc1}$(512,  1)$ and takes  MNIST images
$x$ as  input. ReLU is  used for all  layers except  the last fully  connected layer, which  uses no
activation  function.  We  take  $h$  to  be  the  output  of  $\fc{1}$  and  we  feed  it  to  $G$.
Results can  be found  in \cref{fig:exp:vanilla}.  We plot  the estimate  of the  Wasserstein metric
\cref{eq:wassersteinestimate} against generator iterations in \cref{fig:exp:plot:vanilla}.

\subsection{Gradients  flowing  from fc1  space\label{sec:exp:mnist:hspace}}  We  dispense with  the
ordinary discriminator $D$ in PPGN-$h$ and replace it  with $D_\fc{1}$ attached to \fc{1} of $E$. In
\cref{eq:method:lossgan}  this translates  to $\lambda_{\text{fc1}}=1$  while all  other $\lambda_j$
are  set  to zero.  Note  that  this means  that  the  input of  $G$  and  $D_\fc{1}$ coincide.  The
autoencoder reconstruction loss  \cref{eq:method:aeloss} remains unchanged compared  to the previous
experiment, i.e.  $\alpha_{\fc{1}} =  1$ and  the rest are  $\alpha_j=0$. $D_\fc{1}$  is a  CNN with
fewer parameters  compared to  $D$, \textsf{conv1}$(1,  256, 2)$  $\rightarrow$ \textsf{conv2}$(256,
256,  2)$ $\rightarrow$  \textsf{pool2}  $\rightarrow$ \textsf{conv3}$(256,  512, 2)$  $\rightarrow$
\textsf{pool3} $\rightarrow$  \textsf{disc-fc1}$(512, 1)$.  The use of  activation functions  is the
same as it was for  $D$. We refer to this model as PPGN-$h$-$D_\fc1$ and  train it three times, each
with a  different loss for  $G$. First with  only \lossgan\ to exclude  the effects of  $\lossx$ and
$\lossh$ in order to investigate if $G$ converges using only gradients that flow through $D_\fc{1}$,
see \cref{fig:exp:fc1gan}.  Thereafter, we train  the model with full  loss $L_G$. Samples  for this
experiment  are  found  in  \cref{fig:exp:fc1full}  and  the plot  of  Wasserstein  estimate  is  in
\cref{fig:exp:plot:fc1full}. Lastly, we use only \lossx\ and \lossh\ to check if including \lossgan\
has an impact when training PPGN-$h$-$D_\fc{1}$ with full loss, results in \cref{fig:exp:fc1wo}.

\subsection{Combined approach\label{sec:exp:mnist:combined}} We conduct two experiments that include
both $D$  and $D_\fc{1}$, since  we are interested  in seeing if the  model converges using  the two
discriminators. In the first experiment, we jointly train $G$ together with $D$ and $D_\fc{1}$ using
full loss $L_G$. \cref{fig:exp:combined} contains the results and in \cref{fig:exp:plot:combined} we
find a plot of the Wasserstein estimate \cref{eq:wassersteinestimate}. For short, we name this model
PPGN-$h$-combined. In the second experiment, we randomly select with equal probability only one pair
of adversarial loss  to update the network  with, i.e. either $D$ and  its corresponding adversarial
loss for $G$  or $D_\fc{1}$ and its  adversarial loss.\footnote{Both $D$ and  $D_\fc{1}$ are updated
regardless for the first  $25$ training iterations as well as every $500$th,  in accordance with the
code  accompanied  \cite{arjovsky2017wasserstein}  and  the WGAN-GP  implementation  used  for  this
master's thesis.}  \cref{fig:exp:randomized} shows samples from  the experiment. This last  model is
called PPGN-$h$-random.

\section{Discussion}\label{sec:discussion}   The   evaluation   of   generative   models   is   hard
\cite{theis2015note} and in this section we will  visually compare the results in \cref{sec:exp}. We
realize that this procedure  is subjective, but we feel that it is  appropriate given the simplicity
of the  dataset and limited  scope of the  conclusion we are  about to draw.  We loosely say  that a
dataset is simple if it is not diverse enough and  point to the fact that MNIST images in each class
look somewhat  similar. The  results could  have been  quantified using  the Inception  score method
\cite{salimans2016improved},  but due  to  computational  costs of  using  the  MALA-approx and  the
ease  of  comparing  MNIST  samples  visually  we  omitted  this  step.  In  addition,  recent  work
\cite{barratt2018note}  has  shown  that  using  the  score  for  evaluating  generative  models  is
problematic.

Images produced by  PPGN-$h$-$D_\fc{1}$ \cref{fig:exp:fc1gan} used only $\lossgan$ loss  to take out
the effect of  training $G$ with losses $\lossx$ and  $\lossh$. Here we can see that  $G$ is able to
learn shapes  for all digits.  Clearly, the samples in  \cref{fig:exp:fc1gan} do not  resemble MNIST
digits to  an adequate  degree. Therefore,  we conclude that  it is  not sufficient  to discriminate
between codes $(h, \hat{h})$ using the networks  and hyperparameters we have. However, seeing as the
generator in this case was able to learn  shapes of digits, this prompted us to investigate further.
Subsequently, we  trained the same model  with full loss  $L_G$ which resulted in  improved sampling
quality, \cref{fig:exp:fc1full}. In \cref{fig:exp:plot:fc1full} we see that the Wasserstein estimate
is minimized and flattens  quickly. We hypothesize that this is due  to capacity discrepancy between
$D_\fc{1}$ and  $G$. Alternatively,  $\fc{1}$-space of $E$  is less complex  to minimize  the metric
over, compared to  $X$ space. Furthermore, we cannot  be entirely sure that the  losses $\lossx$ and
$\lossh$ made  $\lossgan$ impractical by interactions  unknown to us,  but it seems likely  that $G$
benefited when  $D_\fc{1}$ was included  given the  results in \cref{fig:exp:fc1gan}.  Therefore, we
trained the same model with only \lossx\ and \lossh\ to see the effects of excluding the adversarial
loss for $G$. The results are shown in \cref{fig:exp:fc1wo} and are somewhat inferior to the samples
in \cref{fig:exp:fc1full}.  We note  that these  models train faster  than Vanilla  PPGN-$h$ because
$D_\fc{1}$ has fewer parameters than $D$. Nevertheless, it can be argued that Vanilla PPGN-$h$ could
be trained with  a slimmer discriminator than  the one we had  designed while the model  at the same
time retains same or  better sampling quality. We did not experiment extensively  with the design of
$D$ and the point raised here should not be dismissed lightly.

Our next set of experiments included both $D$  and $D_\fc{1}$, where we investigate if the generator
$G$  is able  converge  when including  two  different  discriminators. Judging  by  the samples  in
\cref{fig:exp:combined}, we say that  this is the case. Furthermore, the model  was able to minimize
the two different Wasserstein  estimates given by the discriminator respectively, as  can be seen in
\cref{fig:exp:plot:combined}. However, this model  is the most complex in the sense  that it has the
highest number  of learnable parameters and  took longest to  train. Therefore, we trained  the same
model but randomized which discriminator (and its  corresponding adversarial loss for $G$) to update
-- intention here  being to combine the  faster training time of PPGN-$h$-$D_\fc{1}$  and the better
sample quality of Vanilla PPGN-$h$  \cref{fig:exp:vanilla}. The results in \cref{fig:exp:randomized}
are comparable to Vanilla PPGN-$h$.

We refrain from  taking any further conclusion to  the hypothesis, that it is beneficial  for $G$ to
include discriminators  attached to the encoder  $E$, other what has  been said. This is  due to the
simplicity of the  MNIST dataset. To provide more  evidence for the hypothesis we  suggest using the
method proposed here on  more complex datasets and to experiment with  more than two discriminators.
We raise the issue and leave this for future work.

\section{Conclusion}\label{sec:con} In  this master's thesis  we proposed  a method of  training the
Noiseless Joint PPGN-$h$ model  by attaching discriminators to different layers  of the encoder $E$.
We showed that this  approach is viable for the MNIST dataset through  a series of experiments. Yet,
we do not claim that this method generalizes well  for other datasets. The reason is that MNIST is a
rather simple image manifold, compared to ImageNet  and CIFAR-10, and therefore we cannot be certain
that the method works well for more complex manifolds.

\section*{Acknowledgments} I would like to take the  opportunity to show appreciation for family and
friends. Thank you for always being supportive and encouraging.




\footnotesize{
    \bibliographystyle{abbrv}
    \bibliography{refs/gan,refs/ppgn,refs/sets}
}


\end{document}